\documentclass[conference]{IEEEtran}
\IEEEoverridecommandlockouts
\usepackage{cite}
\usepackage{amsmath,amssymb,amsfonts}
\usepackage{algorithmic}
\usepackage{graphicx}
\usepackage{textcomp}

\usepackage[dvipsnames]{xcolor}

\usepackage{tikz}
\usetikzlibrary{positioning}
\usetikzlibrary{shapes.misc}

\usepackage{bm,upgreek}
\usepackage{pgfplots}

\pgfplotsset{compat=1.18} 
\newcommand{\p}{\partial}

\usepackage{hyperref}

\begin{document}

\title{Padding-free Convolution based on Preservation of \\Differential Characteristics of Kernels
}

\author{
\IEEEauthorblockN{Kuangdai Leng}
\IEEEauthorblockA{\textit{Scientific Computing Department} \\
\textit{Science and Technology Facilities Council}\\
Didcot, UK \\
kuangdai.leng@stfc.ac.uk}
\and
\IEEEauthorblockN{Jeyan Thiyagalingam}
\IEEEauthorblockA{\textit{Scientific Computing Department} \\
\textit{Science and Technology Facilities Council}\\
Didcot, UK \\
t.jeyan@stfc.ac.uk}
}

\maketitle

\begin{abstract}
Convolution is a fundamental operation in image processing and machine learning. Aimed primarily at maintaining image size, padding is a key ingredient of convolution, which, however, can introduce undesirable boundary effects. We present a non-padding-based method for size-keeping convolution based on the preservation of differential characteristics of kernels. The main idea is to make convolution over an incomplete sliding window ``collapse'' to a linear differential operator evaluated locally at its central pixel, which no longer requires information from the neighbouring  missing pixels. While the underlying theory is rigorous, our final formula turns out to be simple: the convolution over an incomplete window is achieved by convolving its nearest complete window with a transformed kernel. This formula is computationally lightweight, involving neither interpolation or extrapolation nor restrictions on image and kernel sizes. Our method favours data with smooth boundaries, such as high-resolution images and fields from physics. Our experiments include: i) filtering analytical and non-analytical fields from computational physics and, ii) training convolutional neural networks (CNNs) for the tasks of image classification, semantic segmentation and super-resolution reconstruction. In all these experiments, our method has exhibited visible superiority over the compared ones.  
\end{abstract}


\section{Introduction}
Convolution is a basic operation in image processing. By convolving an image with certain kernels, one can achieve various effects such as blurring, sharpening, and edge detection. The establishment of modern convolutional neural networks (CNNs)~\cite{lecun1998gradient, he2016deep, krizhevsky2017imagenet, huang2017densely} has unprecedentedly highlighted the significance of convolution. CNN-based network architectures have been thriving, such as variational autoencoders (VAEs)~\cite{kingma2013auto}, generative adversarial networks (GANs)~\cite{goodfellow2020generative} and more recently the diffusion models~\cite{ho2020denoising, song2020score}, which spawn numerous derivations for applications. Meanwhile, many techniques have been developed to improve CNNs at a lower level, such as batch normalisation~\cite{ioffe2015batch}, depth-wise separable convolution ~\cite{chollet2017xception}, and skip connections~\cite{he2016deep, huang2017densely}, many of which have become common practice in CNN tasks.

This paper concerns boundary handling, a key ingredient of convolution for maintaining feature map size. Padding is the routine at present. Despite the great success of CNNs with simple padding (e.g., zero padding), previous studies have shown undesirable boundary effects caused by padding, such as artefacts in features~\cite{innamorati2018learning, liu2018partial, liu2021improved} and the spatial bias~\cite{alsallakh2020mind, islam2021position}. A number of techniques have been developed for alleviating padding-induced boundary effects, such as explicit boundary handling (EBH)~\cite{innamorati2018learning}, partial convolution~\cite{liu2018partial}, avoiding uneven application of padding~\cite{alsallakh2020mind}, and quantifiable position-information encoding~\cite{lin2022unveiling}. Though they have been proven or supposed to improve CNNs on different aspects, two major drawbacks persist. First, these techniques only work with back propagation, hence unavailable for image filtering with given kernels. Second, most of these techniques, along with simple padding, are empirically motivated, lacking some rigorous connection between the near-boundary and the interior parts whereby boundary handling can be rendered more interpretable and controllable.

In this paper, we propose a padding-free method for size-keeping convolution, available for both image filtering and CNN training. By introducing a continuous presentation of image over a sliding window, we establish an equivalence between window-wise convolution and pixel-wise differentiation. The latter can be conducted locally at a boundary pixel so that padding is no longer needed. Our final formula is elegantly simple and computationally lightweight: the convolution over an incomplete window is achieved by convolving its nearest complete window with a transformed kernel. 

Boundary handling essentially addresses the issue of missing information, so there cannot exist a method that always prevails for all kinds of data. The reduction from convolution to differentiation makes our method more efficient for images with smoother or more predictable boundaries, such as continuous fields from mathematics and physics, and high-resolution images. Concerning fields, a highly relevant application is CNN-based physics-informed learning~\cite{gao2021phygeonet, gao2021super, shen2021physics, fang2021high, ren2022phycrnet, alguacil2021effects}, aimed at simulating or inverting partial differential equations (PDEs) with physics-embedding loss functions. In a physics-informed CNN, kernels are employed as a discrete representation of differential operators, so preserving the differential characteristics of such kernels at the boundary becomes imperative. First, any artefacts from padding will act as secondary sources to the boundary-value problem, generating fake energies to propagate across the entire domain; second, padding also means imposing a Dirichlet boundary condition~\cite{alguacil2021effects}, which can be incompatible with the given PDE system. Owing to its rich content, we have to discuss physics-informed learning in another paper; here we propose our method for general-purpose image filtering and machine learning.

The remainder of this paper is organised as follows. In the next section, we cover the related work, followed by the description of our method in Section~\ref{sec:method}. We then describe our experiments in Section~\ref{sec:exp}, covering both forward filtering and CNN training. The paper is then concluded in Section~\ref{sec:con}.

\section{Related Work}
\label{sec:related}
Most existing boundary handling techniques are motivated by CNNs, and thus forward-incompatible. We found two methods available for forward image filtering: padding by algebraic extrapolation~\cite{gupta1978note} and a discrete Fourier transformation-based method that involves padding by reflection and circular deconvolution~\cite{aghdasi1996reduction}. We will compare our method with the former, as we could not find a reliable implementation of the latter.

For CNNs, studies have shown that simple padding may not only cause artefacts in features~\cite{innamorati2018learning, liu2018partial, liu2021improved} but also introduce a spatial bias that impairs the translation invariance of CNNs~\cite{alsallakh2020mind, islam2021position, kayhan2020translation, myburgh2020tracking}. Here translation invariance means that CNNs are expected to extract the relevant features regardless of the absolute positions of entities in images. A metric of this bias has recently been proposed in~\cite{lin2022unveiling}. Existing remedies can be largely divided into two categories: advanced padding and relative position encoding, the former motivated more by artefact suppression and the latter by mitigating the spatial bias. Advanced padding includes randomly-valued padding~\cite{nguyen2019distribution}, randomly-positioned padding~\cite{yang2023random}, training an auxiliary CNN for padding~\cite{huang2021context}, symmetric padding with even-sized kernels~\cite{wu2019convolution}, and some non-generic algorithms for domain-specific data~\cite{cheng2018cube, sokouti2016medical, liu2021improved}. Relative position encoding includes eliminating uneven application of padding by constraining the image and kernel sizes~\cite{alsallakh2020mind}, partial convolution~\cite{liu2018partial, liu2022partial}, and EBH~\cite{innamorati2018learning}. Partial convolution is a non-trainable method that first conducts convolution with zero padding and then divides the result by the number fraction of existent pixels in the sliding window. Statistically, it is similar to the randomly-valued padding~\cite{nguyen2019distribution} where the padded values are sampled from a probability distribution determined by a selected boundary vicinity. EBH is the most expensive method, which introduces $(K^2-1)$ duplicates of kernels (where $K$ is the kernel size) to be trained exclusively on the near-boundary pixels grouped by their positions relative to the image boundary. In theory, these $(K^2-1)$ duplicated kernels should maximally reduce boundary effects, but such an extra cost is prohibitively high, especially for a large kernel size (such as $K=7$); besides, these duplicated kernels can only see a small fraction of data near the boundary, so they can converge much slower than the main kernel trained for the bulk interior. Empirically, we do not observe an outstanding advantage of EBH from our CNN experiments.

In Section~\ref{sec:exp}, we will compare our method to the simple padding schemes (zeros, reflect, replicate and circular), along with padding by extrapolation~\cite{gupta1978note}, padding by distribution~\cite{nguyen2019distribution}, partial convolution~\cite{liu2018partial}, and EBH~\cite{innamorati2018learning}.

\section{Method}
\label{sec:method}
We describe our method in this section. Einstein summation convention is adopted for both superscript and subscript indices (in lower case letters) unless they are parenthesised.

\subsection{Forward convolution}
Let $\bm{\upomega}=\{\omega_{ij}$\}, $i, j\in\{0,1,\cdots,K-1\}$ denote the input kernel, which has size $K\times K$, with $K$ being an odd number. Here we assume a square-shaped kernel only to simplify the notations. Consider a pixel ``$\mathtt{a}$'' centring a complete sliding window ``$\mathtt{A}$'', as illustrated in Fig.~\ref{fig:3x3}a, meaning that pixel $\mathtt{a}$ is ``valid'' for convolution. Let $\mathbf{u}^\mathtt{A}=\{u_{ij}^\mathtt{A}\}$, $i, j\in\{0,1,\cdots,K-1\}$ denote the input image given at the $K\times K$ pixels in $\mathtt{A}$ (after stride and dilation if required). The convolution over $\mathtt{A}$ can then be written as
\begin{equation}
z^\mathtt{a}:=\omega_{ij}u_{ij}^\mathtt{A}=\bm{\upomega}:\mathbf{u}^\mathtt{A}.
\label{eq:zqconv}
\end{equation}
We aim for a non-padding-based method to accomplish such convolution at the ``invalid'' pixels that centre incomplete sliding windows, such as ``$\mathtt{b}$'' centring ``$\mathtt{B}$'' in Fig.~\ref{fig:3x3}a.

A continuous, sub-pixel image can be formed in $\mathtt{A}$ using the Lagrange interpolating polynomial, as denoted by $\tilde{u}^\mathtt{A}(h,w)$ with $h$ and $w$ being the spatial coordinates:
\begin{equation}
\tilde{u}^\mathtt{A}(h,w):=l_i(h) l_j(w) u_{ij}^\mathtt{A},\quad  h,w\in [0, K-1],
\label{eq:interp}
\end{equation}
where $l_i(x)$ is the Lagrange basis simplified for a uniform grid with a unit interval,
\begin{equation}
l_i(x) = \prod_{\substack{0\le k \le K-1\\k\ne i}} \frac{x-k}{i-k}, \quad  i\in \{0,1,\cdots ,K-1\}.
\end{equation}
The interpolation in eq.~\eqref{eq:interp} leads to a 2D polynomial of degree $(K-1)$ that preserves the pixel values, i.e., $\tilde{u}^\mathtt{A}(i,j)=u^\mathtt{A}_{ij}$. Spatial derivatives of $\tilde{u}^\mathtt{A}(h,w)$ can then be conducted up to order $(K-1)$ in each direction, formulated as
\begin{equation}
\left.\mathcal{D}^{mn}\tilde{u}^\mathtt{A}\right|_{(h,w)}:=
\left.\frac{\p^{(m+n)} \tilde{u}^\mathtt{A}}{\p h^{(m)} \p w^{(n)}}\right|_{(h,w)}
=l_i^{m}(h) l_j^{n}(w) u_{ij}^\mathtt{A},
\label{eq:DuQhw}
\end{equation}
for $m, n\in \{0, 1, \cdots, K-1\}$, where $l_i^{m}(x)$ is the \mbox{$m$-th} derivative of $l_i(x)$, available in exact form given $K$ (as they are all rational numbers), and $\mathcal{D}^{mn}$ denotes the partial differential operator of order $(m,n)$, e.g., $\mathcal{D}^{12}=\frac{\p^3 }{\p h \p w^2}$. Evaluated at a pixel with location $(p,q)$, for $p,q\in\{0,1,\cdots,K-1\}$, eq.~\eqref{eq:DuQhw} yields
\begin{equation}
\left.\mathcal{D}^{mn}\tilde{u}^\mathtt{A}\right|_{(p,q)}=\delta_{pq,ij}^{mn} u_{ij}^\mathtt{A}=
\bm{\updelta}_{pq}^{mn} :\mathbf{u}^\mathtt{A},
\label{eq:DuQpq}
\end{equation}
based on the definition that
\begin{equation}
    \delta_{pq,ij}^{mn}:=l_i^{m}(p) l_j^{n}(q).
    \label{eq:defdelta}
\end{equation}
Equation~\eqref{eq:DuQpq} states that, $\left.\mathcal{D}^{mn}\tilde{u}^\mathtt{A}\right|_{(p,q)}$, the $(m,n)$-th order derivative of our continuous image evaluated at pixel $(p,q)$, can be obtained by convolving the image $\mathbf{u}^\mathtt{A}$ with the above-defined kernel $\bm{\updelta}^{mn}_{pq}$. Therefore, we call $\bm{\updelta}$ defined by eq.~\eqref{eq:defdelta} the \emph{differential kernels}, which depends only on the kernel size $K$. An example for $K=3$ is provided in Fig.~\ref{fig:3x3}b.

\begin{figure*}
\centering
\begin{tikzpicture}[node distance=.4cm]
\node[{circle, draw=Green, fill=Green!15, very thick}](A11){$u_{11}^\mathtt{A}$};
\node[{circle, draw=Black, very thick}](A12)[right=of A11]{$u_{12}^\mathtt{A}$};
\node[{circle, draw=Black, very thick}](A21)[below=of A11]{$u_{21}^\mathtt{A}$};
\node[{circle, draw=Black, very thick}](A22)[below=of A12]{$u_{22}^\mathtt{A}$};
\node[{rounded corners=0.2cm, draw=Black, very thick,minimum size=.9cm}](A01)[above=of A11]{$u_{01}^\mathtt{A}$};
\node[{rounded corners=0.2cm, draw=Black, very thick,minimum size=.9cm}](A02)[above=of A12]{$u_{02}^\mathtt{A}$};
\node[{rounded corners=0.2cm, draw=Black, very thick,minimum size=.9cm}](A20)[left=of A21]{$u_{20}^\mathtt{A}$};
\node[{rounded corners=0.2cm, draw=Black, very thick,minimum size=.9cm}](A10)[left=of A11]{$u_{10}^\mathtt{A}$};
\node[{rounded corners=0.2cm, draw=Red, fill=Red!15, very thick,minimum size=.9cm}](A00)[left=of A01]{$u_{00}^\mathtt{A}$};
\node[](pixela)[above=-.09cm of A11]{\color{Green} Pixel $\mathtt{a}$};
\node[](pixelb)[above=-.09cm of A00]{\color{Red} Pixel $\mathtt{b}$};
\node[{rounded corners=0.2cm, draw=Black, very thick,minimum size=.9cm}](A03)[right=of A02]{};
\node[{rounded corners=0.2cm, draw=Black, very thick,minimum size=.9cm}](A13)[right=of A12]{};
\node[{rounded corners=0.2cm, draw=Black, very thick,minimum size=.9cm}](A23)[right=of A22]{};
\node[{rounded corners=0.2cm, draw=Black, very thick,minimum size=.9cm}](A30)[below=of A20]{};
\node[{rounded corners=0.2cm, draw=Black, very thick,minimum size=.9cm}](A31)[below=of A21]{};
\node[{rounded corners=0.2cm, draw=Black, very thick,minimum size=.9cm}](A32)[below=of A22]{};
\node[{rounded corners=0.2cm, draw=Black, very thick,minimum size=.9cm}](A33)[below=of A23]{};
\node at (A11.center) [{draw=Green, thick, minimum size=4.3cm}](WinA){};
\node at (A00.center) [{draw=Red, thick, minimum size=4.3cm}](WinB){};
\node[{draw=Black, cross out, very thick,minimum size=.6cm}](X)[left=.65cm of A00]{};
\node[{draw=Black, cross out, very thick,minimum size=.6cm}][left=.65cm of A10]{};
\node[{draw=Black, cross out, very thick,minimum size=.6cm}][above=.65cm of A00]{};
\node[{draw=Black, cross out, very thick,minimum size=.6cm}](X01)[above=.65cm of A01]{};
\node[{draw=Black, cross out, very thick,minimum size=.6cm}](X)[above=.8cm of X]{};
\node[{rounded corners=0.2cm, very thick,minimum size=.9cm}](WA)[above=of A02, xshift=.3cm, yshift=-.3cm]{\color{Green} Window $\mathtt{A}$};
\node[{very thick,minimum size=.9cm}](WB)[above=of WA, yshift=-.5cm]{\color{Red} Window $\mathtt{B}$};

\node[{very thick,minimum size=.9cm}](setup)[above=of WinB.north west, anchor=west, yshift=0cm]{(a) Image setup};

\node[{circle, draw=Black, very thick,minimum size=.4cm}](C)[below=of X,xshift=-.5cm, yshift=-5.7cm]{};
\node[][right=of C, xshift=-.3cm](CT){Pixels centring a complete sliding window};
\node[{rounded corners=0.1cm, draw=Black, very thick,minimum size=.4cm}](R)[below=of C, yshift=.1cm]{};
\node[][right=of R, xshift=-.3cm](RT){Pixels centring an incomplete sliding window};
\node[{cross out, draw=Black, very thick,minimum size=.3cm, yshift=.1cm}](XX)[below=of R]{};
\node[][right=of XX, xshift=-.3cm](XT){Absent pixels outside image boundary};

\node[][right=of setup, xshift=5cm](diff)
{(b) Differential kernels at central pixel  $\mathtt{a}$};

\node[][fill=Black!20, below=of diff,  xshift=-1.35cm, yshift=-2.5cm, minimum width=3.3cm, minimum height=1.4cm](shader){};

\node[][below=of diff.west, anchor=north west, yshift=.1cm](d0011){
$
\begin{aligned}
&\begin{aligned}
&\bm{\updelta}^{00}_{11}
=\begin{pmatrix}
0&0&0\\
0&1&0\\
0&0&0
\end{pmatrix} 
&\Leftarrow&\quad \mathcal{D}^{00}=1\\
&\bm{\updelta}^{01}_{11}
=\dfrac{1}{2}\begin{pmatrix}
0&0&0\\
-1&0&1\\
0&0&0
\end{pmatrix} 
&\Leftarrow& \quad\mathcal{D}^{01}=\dfrac{\p}{\p w}\\
&\bm{\updelta}^{02}_{11}
=\begin{pmatrix}
0&0&0\\
1&-2&1\\
0&0&0
\end{pmatrix} 
&\Leftarrow& \quad\mathcal{D}^{02}=\dfrac{\p^2}{\p w^2}\\
&\bm{\updelta}^{11}_{11}
=\dfrac{1}{4}\begin{pmatrix}
1&0&-1\\
0&0&0\\
-1&0&1
\end{pmatrix} 
&\Leftarrow& \quad\mathcal{D}^{11}=\dfrac{\p}{\p h} \dfrac{\p}{\p w}\\
&\bm{\updelta}^{12}_{11}
=\dfrac{1}{2}\begin{pmatrix}
-1&2&-1\\
0&0&0\\
1&-2&1
\end{pmatrix} 
&\Leftarrow& \quad\mathcal{D}^{12}=\dfrac{\p}{\p h} \dfrac{\p^2}{\p w^2}\\
&\bm{\updelta}^{22}_{11}
=\begin{pmatrix}
1&-2&1\\
-2&4&-2\\
1&-2&1
\end{pmatrix} 
&\Leftarrow& \quad\mathcal{D}^{22}=\dfrac{\p^2}{\p h^2} \dfrac{\p^2}{\p w^2}\\
% \bm{\updelta}^{10}_{11}
% =\left(\bm{\updelta}^{01}_{11}\right)^\top,\quad
% \bm{\updelta}^{20}_{11}
% =\left(\bm{\updelta}^{02}_{11}\right)^\top,\quad
% \bm{\updelta}^{21}_{11}
% =\left(\bm{\updelta}^{12}_{11}\right)^\top
\end{aligned}
\\
&\bm{\updelta}^{10}_{11}
=\left(\bm{\updelta}^{01}_{11}\right)^\top,\quad
\bm{\updelta}^{20}_{11}
=\left(\bm{\updelta}^{02}_{11}\right)^\top,\quad
\bm{\updelta}^{21}_{11}
=\left(\bm{\updelta}^{12}_{11}\right)^\top
\end{aligned}
$
};
% \node[][below=of d0011.south west, anchor=north west,yshift=.2cm](trans){
% $
% $
% };

\node[][below=of XX.west,  xshift=-.2cm, yshift=-.3cm, anchor=west](assemble){(c) Assembled differential kernels: $\vec{\mathbf{D}}_{11}$ for central pixel $\mathtt{a}$ and $\vec{\mathbf{D}}_{00}$ for target pixel $\mathtt{b}$ ($R=S=0$)
};

\node[][fill=Black!20, below=of assemble.west,  xshift=2.85cm, yshift=-.0cm, minimum width=0.7cm, minimum height=3.8cm](shader){};

\node[][below=of assemble.west, anchor=north west, yshift=0.1cm](DMM){
$
\setlength\arraycolsep{3pt}
\vec{\mathbf{D}}_{11}=
\begin{pmatrix}
0 & 0 & 0 & 0 & \frac{1}{4} & -\frac{1}{2} & 0 & -\frac{1}{2} & 1
\\
0 & 0 & 0 & -\frac{1}{2} & 0 & 1 & 1 & 0
& -2\\
0 & 0 & 0 & 0 & -\frac{1}{4} & -\frac{1}{2} & 0 & \frac{1}{2} & 1
\\
0 & -\frac{1}{2} & 1 & 0 & 0 & 0 & 0 & 1 & -2
\\
1 & 0 & -2 & 0 & 0 & 0 & -2 & 0 & -4
\\
0 & \frac{1}{2} & 1 & 0 & 0 & 0 & 0 & -1 & -2
\\
0 & 0 & 0 & 0 & -\frac{1}{4} & \frac{1}{2} & 0 & -\frac{1}{2} & 1
\\
0 & 0 & 0 & \frac{1}{2} & 0 & -1 & 1 & 0 & -2
\\
0 & 0 & 0 & 0 & \frac{1}{4} & \frac{1}{2} & 0 & \frac{1}{2} & 1
\\
\end{pmatrix},\quad
\vec{\mathbf{D}}_{00}=
\begin{pmatrix}
    1&-\frac{3}{2}&1&-\frac{3}{2}&\frac{9}{4}&-\frac{3}{2}&1&-\frac{3}{2}&1\\
    0&2&-2&0&-3&3&0&2&-2\\
    0&-\frac{1}{2}&1&0&\frac{3}{4}&-\frac{3}{2}&0&-\frac{1}{2}&1\\
    0&0&0&2&-3&2&-2&3&-2\\
    0&0&0&0&4&-4&0&-4&4\\
    0&0&0&0&-1&2&0&1&-2\\
    0&0&0&-\frac{1}{2}&\frac{3}{4}&-\frac{1}{2}&1&-\frac{3}{2}&1\\
    0&0&0&0&-1&1&0&2&-2\\
    0&0&0&0&\frac{1}{4}&-\frac{1}{2}&0&-\frac{1}{2}&1
\end{pmatrix}
$
};

\node[][below=of XX.south west,  xshift=-.2cm, yshift=-4.8cm, anchor=west](ker){(d) Input kernel $\bm{\upomega}$ (a box-blur filter) and the transformed kernel
$\bm{\upvarpi}$};

\node[][below=of ker.west, anchor=north west, yshift=.1cm](omega){
$
\bm{\upomega}=
\begin{pmatrix}
1 & 1 & 1\\
1 & 1 & 1\\
1 & 1 & 1\\
\end{pmatrix}
\quad
\Longrightarrow
\quad
\bm{\upvarpi}=
\begin{pmatrix}
16 & -8 & 4\\
-8 & 4 & -2\\
4 & -2 & 1\\
\end{pmatrix},\quad\text{by}
\ \ 
\vec{\bm{\upvarpi}}=\vec{\mathbf{D}}_{00}\cdot\vec{\mathbf{D}}_{11}^{-1}\cdot\vec{\bm{\upomega}}
$
};

\end{tikzpicture}
\vspace{-.2cm}
\caption{A complete example of our method: convolving a $4\times 4$ image with a $3\times 3$ kernel. We aim to conduct convolution at invalid pixel $\mathtt{b}$, whose nearest complete window is $\mathtt{A}$. Based on eq.~\eqref{eq:ostar}, the outcome is given by $z^{\mathtt{b}}=\bm{\upvarpi}:\mathbf{u}^\mathtt{A}$, where $\vec{\bm{\upvarpi}}=\vec{\mathbf{D}}_{00}\cdot\vec{\mathbf{D}}_{11}^{-1}\cdot\vec{\bm{\upomega}}$, as the location of $\mathtt{b}$ in $\mathtt{A}$ is $(R,S)=(0,0)$ and $M=\frac{K-1}{2}=1$. The two shaded areas in (b) and (c) mark out that $\bm{\updelta}^{02}_{11}$ constructs the third column of $\vec{\mathbf{D}}_{11}$, according to eq.~\eqref{eq:vec}. The $\vec{\mathbf{D}}$ metrics and their inverse are precomputed.
}
\label{fig:3x3}
\end{figure*}

Our central idea is to \emph{represent the wanted convolution or eq.~\eqref{eq:zqconv} over a generic window $\mathtt{B}$, complete or incomplete, as a unique linear differential operator $\mathcal{L}$ applied on the continuous image $\tilde{u}(h,w)$ and evaluated \textbf{locally} at the centre $\mathtt{b}$}. Formally, we prescribe the following equivalence:
\begin{equation}
\bm{\upomega}:\mathbf{u}^\mathtt{B}\equiv
\left.\mathcal{L}\tilde{u}\right|_{\mathtt{b}},\quad \mathcal{L}:=\alpha^{mn}\mathcal{D}^{mn},
\label{eq:equiv}
\end{equation}
where $\alpha^{mn}$, for $m,n\in\{0,1,\cdots,K-1\}$, are the real coefficients of the linear differential operator $\mathcal{L}$. It must be emphasised that eq.~\eqref{eq:equiv} does not specify how $\tilde{u}(h,w)$ is determined, implying that the window for its interpolation does not need to be our target window $\mathtt{B}$. This will eventually enable convolution over the incomplete windows.

The coefficients $\alpha^{mn}$ are determined such that eq.~\eqref{eq:equiv} holds \emph{at every valid pixel} given $\tilde{u}(h,w)$ interpolated by its centred window. A generic example is our pixel $\mathtt{a}$ that centres window $\mathtt{A}$. The local coordinates of $\mathtt{a}$ in $\mathtt{A}$ are $(M,M)$, with $M=\frac{K-1}{2}$. With $\tilde{u}(h,w)$ interpolated by $\mathtt{A}$, eq.~\eqref{eq:equiv} becomes 
\begin{equation}
\begin{aligned}
\bm{\upomega}:\mathbf{u}^\mathtt{A}&\equiv
\left.\mathcal{L}\tilde{u}\right|_\mathtt{a}\equiv
\left.\mathcal{L}\tilde{u}^\mathtt{A}\right|_\mathtt{a}\\&=\left.\alpha^{mn}\mathcal{D}^{mn}\tilde{u}^\mathtt{A}\right|_{(M,M)}=\alpha^{mn}\bm{\updelta}_{MM}^{mn}:\mathbf{u}^\mathtt{A},
\label{eq:equivvalid}
\end{aligned}
\end{equation}
the last part using our definition of $\bm{\updelta}^{mn}_{pq}$ in eq.~\eqref{eq:defdelta}. To make eq.~\eqref{eq:equivvalid} hold regardless of data $\mathbf{u}^\mathtt{A}$, $\alpha^{mn}$ must be the solution of the following $K^2\times K^2$ linear system:
\begin{equation}
\alpha^{mn}\bm{\updelta}_{MM}^{mn}=\bm{\upomega}.
\label{eq:linear}
\end{equation}
If we denote the vectorisation of a generic $K\times K$ matrix $\mathbf{a}$ by $\vec{\mathbf{a}}$ such that $\vec{a}_{i K+j}=a_{ij}$, the above linear system can be recast to the following standard form:
\begin{equation*}
\renewcommand\arraystretch{1.7} 
\setlength\arraycolsep{2pt}
\begin{aligned}
&\underbrace{
\begin{pmatrix}
\vec{\delta}_{MM,0}^{0} & \vec{\delta}_{MM,0}^{1} &
\cdots &
\vec{\delta}_{MM,0}^{K^2-1} &
\\
\vec{\delta}_{MM,1}^{0} & \vec{\delta}_{MM,1}^{1} &
\cdots &
\vec{\delta}_{MM,1}^{K^2-1} &
\\
\vdots &  \vdots & \ddots & \vdots &
\\
\vec{\delta}_{MM,K^2-1}^{0} & \vec{\delta}_{MM,K^2-1}^{1} &
\cdots &
\vec{\delta}_{MM,K^2-1}^{K^2-1} &
\end{pmatrix}
}_{
\begin{matrix}
\vec{\mathbf{D}}_{MM}
\end{matrix}
}
\underbrace{
\begin{pmatrix}
\vec{\alpha}^0\\
\vec{\alpha}^1\\
\vdots\\
\vec{\alpha}^{K^2-1}
\end{pmatrix}
}_{
\begin{matrix}
\vec{\bm{\upalpha}}
\end{matrix}
}
\end{aligned}
\end{equation*}
\vspace{-.4cm}
\begin{flalign}
\renewcommand\arraystretch{1.7} 
\setlength\arraycolsep{2pt}
&\begin{aligned}
=
\underbrace{
\begin{pmatrix}
\vec{\omega}_0&
\vec{\omega}_1&
\cdots&
\vec{\omega}_{K^2-1}
\end{pmatrix}^\top
}_{
\begin{matrix}
\vec{\bm{\upomega}}^\top
\end{matrix}
}.
\label{eq:vec}
\end{aligned}&&
\end{flalign}
The assembled matrix $\vec{\mathbf{D}}_{MM}$ is always invertible because the $K^2$ differential kernels, $\vec{\bm{\updelta}}^{k}_{MM}$ for $k\in \{0, 1, \cdots, K^2-1\}$ (i.e., each column in $\vec{\mathbf{D}}_{MM}$), are linearly independent given that $l_i(x)$ is a complete polynomial of degree $(K-1)$. The inverse of $\vec{\mathbf{D}}_{MM}$ can be exactly shown for a given kernel size $K$, so computing $\vec{\bm{\upalpha}}=\vec{\mathbf{D}}_{MM}^{-1}\cdot\vec{\bm{\upomega}}$ is trivial.

Having uniquely determined the differential operator $\mathcal{L}$ as $\alpha^{mn}\mathcal{D}^{mn}$, we can evaluate $\mathcal{L}$ at any invalid pixel once a continuous image in its neighbourhood is provided. Here we choose to determine such a continuous image by its nearest complete window. Let $\mathtt{b}$ be an invalid pixel centring an incomplete window $\mathtt{B}$, and $\mathtt{A}$ be the complete window nearest to $\mathtt{b}$, such as Fig.~\ref{fig:3x3}a. It is straightforward to show that $\mathtt{b}$ lies in $\mathtt{A}$ (but not at its centre). Assume that the local coordinates of $\mathtt{b}$ in $\mathtt{A}$ are $(R,S)$, with $R,S\in\{0,1,\cdots,K-1\}$ and $(R,S)\ne(M,M)$. We use the R.H.S. of eq.~\eqref{eq:equiv} to compute the convolution over $\mathtt{B}$, however, with the continuous image $\tilde{u}(h,w)$ interpolated from $\mathtt{A}$:
\begin{equation}
\begin{aligned}
\bm{\upomega}:{\color{red}{\mathbf{u}^\mathtt{B}}}\equiv&
\left.\mathcal{L}\tilde{u}\right|_\mathtt{b}\approx
\left.\mathcal{L}\tilde{u}^\mathtt{A}\right|_\mathtt{b}\\=&\left.\alpha^{mn}\mathcal{D}^{mn}\tilde{u}^\mathtt{A}\right|_{(R,S)}=\alpha^{mn}\bm{\updelta}_{RS}^{mn}:\mathbf{u}^\mathtt{A}.
\label{eq:invalid}
\end{aligned}
\end{equation}
We colour $\mathbf{u}^\mathtt{B}$ in red to indicate that it is partially unavailable. Note that the ``$\approx$'' sign in the above equation indicates the only approximation we have introduced: the window by which the continuous image $\left.\tilde{u}\right|_\mathtt{b}$ is interpolated. More readably, using $\vec{\bm{\upalpha}}=\vec{\mathbf{D}}_{MM}^{-1}\cdot\vec{\bm{\upomega}}$, eq.~\eqref{eq:invalid} can be simplified as
\begin{equation}
\vec{\bm{\upomega}}\cdot {\color{red}\vec{\mathbf{u}}^\mathtt{B}}\approx
\vec{\bm{\upvarpi}}\cdot\vec{\mathbf{u}}^\mathtt{A}, \quad 
\vec{\bm{\upvarpi}}:=\vec{\mathbf{D}}_{RS}\cdot\vec{\mathbf{D}}_{MM}^{-1}\cdot\vec{\bm{\upomega}}.
\label{eq:ostar}
\end{equation}
We refer to eq.~\eqref{eq:ostar} as the \emph{differential kernel transformation}. Clearly, it is compatible with the valid pixels, for which $R=S=M$ and thus ${\bm{\upvarpi}}={\bm{\upomega}}$. Refer to Fig.~\ref{fig:3x3} for the example of $K=3$. In summary, \emph{the convolution with $\bm{\upomega}$ over an incomplete window $\mathtt{B}$ is conducted by a ``shifted'' convolution with the transformed kernel $\bm{\upvarpi}$ over its nearest complete window $\mathtt{A}$, under a window shift of $(M-R,M-S)$}.

\subsection{Method properties}
Our method has the following key properties:

\subsubsection{Theoretical soundness}
Our method preserves the differential characteristics of kernels at the invalid pixels by making the window-wise convolution collapse to a pixel-wise differential operator, or eq.~\eqref{eq:equiv}. Such a connection between the ``invalid'' near-boundary part and the ``valid'' interior part is theoretically sound and self-contained. In contrast, most previous methods are empirically motivated by CNNs.

\subsubsection{Only using original pixel values}
Though we have introduced a continuous image conceptually, our final formula for convolution at the invalid pixels, eq.~\eqref{eq:ostar}, operates on the original pixels from the input image. Without introducing extra information outside the image boundary (such as by padding or extrapolation) or between pixels (such as by interpolation), our method avoids these sources of artefacts. This property also makes our method compatible with stride and dilation for CNN training.

\subsubsection{Low overhead} 
The transformation matrices in eq.~\eqref{eq:ostar}, $\vec{\mathbf{D}}_{RS}\cdot\vec{\mathbf{D}}_{MM}^{-1}$, are constants given the kernel size $K$, which can be precomputed for frequently-used $K$'s, such as 3, 5 and 7. Therefore, eq.~\eqref{eq:ostar} demands a low computational cost in both forward and back propagation.

\subsubsection{Favouring data with smooth boundaries} 
Let $\tilde{u}^*(h,w)$ denote the true image function, which can be discontinuous or even inexpressible. At the boundary pixels, our method preserves the convolution-associated differential operator up to order $(K-1)$. Therefore, it becomes exact when the local Taylor expansion of $\tilde{u}^*(h,w)$ on the boundary is of order $(K-1)$ or lower, i.e., when $\tilde{u}^*(h,w)$ is sufficiently smooth on the boundary. The error of our method increases as $\tilde{u}^*(h,w)$ becomes more non-smooth or unpredictable near the boundary. Datasets underpinned by physical or mathematical processes, such as solutions of partial differential equations and tomographic images, tend to benefit from our method. Furthermore, our method tends to work better with higher-resolution images because $\tilde{u}^*(h,w)$ becomes smoother as the sliding windows shrink with respect to image contents.

\section{Experiments}
\label{sec:exp}
We evaluate our method with two types of experiments, the former on image filtering with given kernels, as reported in Section~\ref{sec:filter}, and the latter on CNN-based computer vision tasks, as reported in Section~\ref{sec:cnn}. Einstein
summation convention is not used in this section.

\subsection{Image filtering}
\label{sec:filter}
\begin{figure*}
    \centering
    \includegraphics[width=\textwidth]{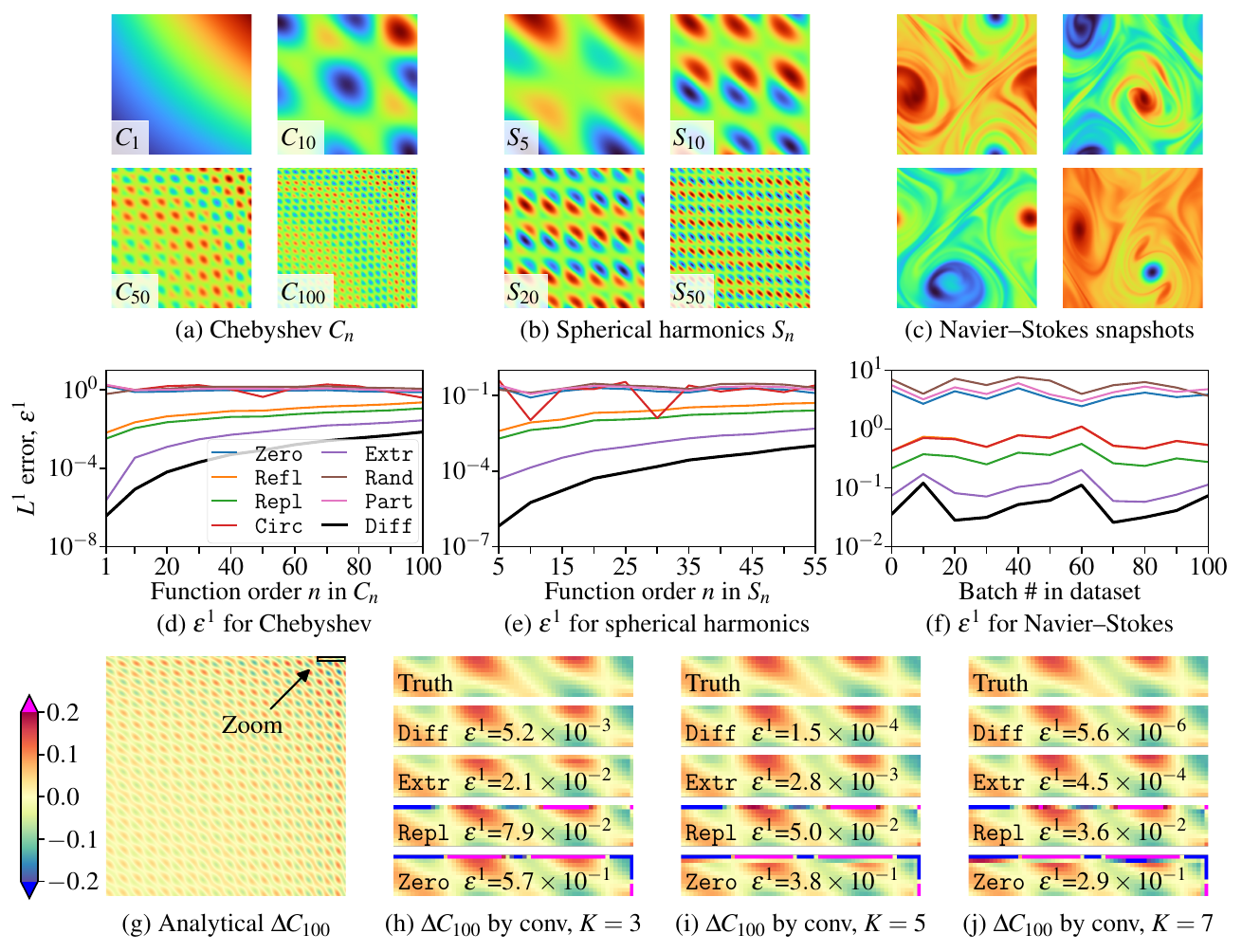}
    \vspace{-.7cm}
    \caption{Filtering three datasets by convolution: Chebyshev, spherical harmonics and solutions of Navier-Stokes equations~\cite{li2021physics}. The top row shows our target functions. The second row shows the $L^1$ errors ($\varepsilon^1$) for 100 random filters applied to the datasets. We vary the function order $n$ for Chebyshev and spherical harmonics, and the batch index for the Navier-Stokes.
    In the bottom row, we apply the Laplace filters ($\Delta=\partial_{hh}+\partial_{ww}$) of sizes 3, 5 and 7 to $C_{100}$; (g) shows the ground truth of $\Delta C_{100}$ based on analytical evaluation, and (h)$\sim$(j) the results by convolution, zooming into the annotated box in (g).  
    }
    \label{fig:filter}
\end{figure*}

We consider three synthetic datasets. The first two are analytical 2D functions, respectively generated from the Chebyshev polynomials and the spherical harmonics. Both are popular basis functions in computational physics and mathematics. Therefore, the accuracy of our method as tested on these basis functions can reasonably indicate its versatility for handling extensive continuous fields. Our Chebyshev-based functions are given by 
\begin{equation}
C_n(h,w)=U_{n}(h)U_{n}(w)\sin\left(n(h+w)\right),    
\end{equation}
where the $U_{n}$'s, for $n\in\{0,1,2,\cdots\}$, are the Chebyshev polynomials of the second kind, and the rotation by $\sin\left(n(h+w)\right)$ makes the 2D patterns non-parallel to the axes. The spherical harmonic-based functions are given by
\begin{equation}
S_n(h,w)=Y_{2n}^{n}(h, w)\sin\left(n(h+w)\right), 
\end{equation}
where $Y_{l}^{m}$ is the spherical harmonic function of degree $l$ and order $m$, composed of trigonometric functions and the associated Legendre polynomials.
As the function order $n$ grows, both $C_{n}$ and $S_{n}$ becomes more oscillating or non-smooth, as shown in Fig.~\ref{fig:filter}a and \ref{fig:filter}b. 
Our third dataset is non-analytical, including numerical solutions of the Navier–Stokes equations for turbulence, as shown in \ref{fig:filter}c, borrowed from the physics-informed neural operators~\cite{li2021physics}.

We apply random filters to these datasets and compare the accuracy of the following eight methods: padding respectively by zeros (\texttt{Zero}), reflection (\texttt{Refl}), replicate (\texttt{Repl}), circular (\texttt{Circ}), extrapolation (\texttt{Extr})~\cite{gupta1978note}, and distribution (\texttt{Rand})~\cite{nguyen2019distribution}, along with partial convolution (\texttt{Part})~\cite{liu2018partial} and our differentiation-based method (\texttt{Diff}). For \texttt{Extr}, we use linear, quadratic and cubic respectively for $K=3, 5$ and $7$. For \texttt{Rand}, the padded values are sampled from the four normal distributions determined respectively for the top, bottom, left and right edges with a thickness of $(K+1)/2$. Requiring back-propagation, EBH~\cite{innamorati2018learning} is not applicable here.

The superiority of our method against the others is visible from Fig.~\ref{fig:filter}. The second row shows the $L^1$ errors of each method for 100 random filters applied to the three datasets. Our method proves to be remarkably more accurate than the others (note that the y-axes are logarithmic). For Chebyshev and spherical harmonics, the errors increase with the function order $n$, but our method prevails across all the orders.
The bottom row of Fig.~\ref{fig:filter} zooms into the boundary artefacts caused by the Laplace filters with different kernel sizes. It is shown that our method is visually artefact-free even at $n=100$; \texttt{Extr} also works reasonably well, but its induced errors are still visible and 1$\sim$2 orders of magnitude larger than ours. 
The other padding schemes and partial convolution will cause strong artefacts irrespective of kernel size and function order, so they are unsuitable for the task of image filtering.

\subsection{Learning with CNNs}
\label{sec:cnn}
In this section, we consider three common CNN tasks: image classification, semantic segmentation and super-resolution reconstruction. For all these experiments we use real-world datasets (instead of e.g. analytical fields) to avoid favouritism towards our method by means of forward filtering.
The original U-Net architecture \cite{ronneberger2015u} is adopted, with its \texttt{Conv2d} layers varying among nine different boundary handling methods.
The first eight are those from the previous experiment: \texttt{Zero}, \texttt{Refl}, \texttt{Repl}, \texttt{Circ}, \texttt{Extr}, \texttt{Rand}, \texttt{Part} and \texttt{Diff} (ours), all non-trainable, and the last one is \texttt{EBH}~\cite{innamorati2018learning}, involving eight duplicates of kernels (as $K=3$ in a U-Net). 
The relative wall-times for training the nine U-Nets are reported in Table~\ref{tab:metrics}, which shows that our method runs as fast as circular padding (PyTorch built-in).
For each problem, the nine U-Nets are initialised with the same weights, and we loop over five random seeds to obtain the reported metric scores. 
It must be emphasised that \emph{boundary handling is a low-level operation in CNNs, so we use a simple network architecture and loss functions to isolate its influences} instead of pursuing the state of the art of the considered problems (datasets) with any advanced yet irrelevant techniques.

\subsubsection{Classification}
\label{sec:class}
We use the Caltech-101 dataset~\cite{fei2006one} for this experiment. The latent (bottom layer) of the U-Net is connected to a two-layer fully-connected network to predict the soft labels and then the classification (cross-entropy) loss. The total loss is the sum of the classification loss and 1\% of the reconstruction loss (which accelerates convergence). The images are all reshaped to $448\times448$.

The accuracy of the models on the test set (20\% of data) is reported in Table~\ref{tab:metrics}. 
It is shown that \texttt{Part}, \texttt{EBH} and \texttt{Diff} have achieved a much higher accuracy ($>50\%$) than the other six padding-based methods ($<30\%$), with our \texttt{Diff} attaining the highest. 
The padding-based U-Nets have mostly failed to learn (with many hyperparameters tested), as can be seen from their low accuracy and training history (not shown here for brevity). 
This can be a good example of padding-free boundary handling (\texttt{Part}, \texttt{EBH} and \texttt{Diff}) significantly enhancing the learnability of a CNN.

\begin{table*}
\caption{Metric Scores for the CNN-based Experiments}
\begin{center}
\renewcommand{\arraystretch}{1.3}
\begin{tabular}{|c|c|c|c|c|c|c|c|c|c|}
\hline
  & \textbf{Classification}  & \multicolumn{3}{|c}{\textbf{Semantic Segmentation}} & \multicolumn{3}{|c}{\textbf{Super-resolution Reconstruction}} & \multicolumn{2}{|c|}{\textbf{Computational cost}} 
  \\
   \textbf{Method} & \textbf{Caltech-101}  & \multicolumn{3}{|c}{\textbf{Cityscapes}} & \multicolumn{3}{|c}{\textbf{ETOPO-15}$''$} & \multicolumn{2}{|c|}{\textbf{Original U-Net}} \\
\cline{2-10}

 & 
\textbf{\emph{Accuracy}}$^\mathrm{a}$& 
\textbf{\emph{IoU}} & 
\textbf{\emph{F1}} & 
\textbf{\emph{Accuracy}} & 
${\varepsilon^2_\text{INTER} \times 10^4}$ & 
${\varepsilon^2_\text{FRAME} \times 10^4}$ & 
${\varepsilon^2_\text{FRAME} / \varepsilon^2_\text{INTER}}$ & \textbf{\emph{\# kernels}}$^\mathrm{b}$ & \textbf{\emph{Wall-time}}$^\mathrm{c}$
 \\

\hline

\texttt{Zero} & 30.7\% & 
84.8\% & 91.8\% & 97.9\% &
7.0 & 10.2 &  144.5\% & 1 & 1.0 \\

\texttt{Refl} & 28.8\% & 
84.6\% & 91.6\% & 97.7\% &
7.2 & 10.0 & 139.9\% & 1 & $\sim$1.1\\

\texttt{Repl} & 22.9\% & 
85.1\% & 92.0\% & 97.9\% &
7.0 & \emph{\textbf{9.2}} & \emph{\textbf{132.0\%}} & 1 & $\sim$1.1\\

\texttt{Circ} & 29.1\% & 85.0\% & 91.9\% & 97.9\% &
7.2 & 11.6 & 157.9\% & 1 & $\sim$1.5\\

\texttt{Extr}~\cite{gupta1978note} & 28.8\% & 
85.3\% & 92.1\% & 98.0\% &
7.4 & 10.9& 147.5\% & 1 & $\sim$5.8\\

\texttt{Rand}~\cite{nguyen2019distribution} & 27.9\% & 
85.1\% & 91.9\% & 97.9\% &
7.7 & 15.5 & 200.2\% & 1 & $\sim$2.6\\

\texttt{Part}~\cite{liu2018partial} & \emph{\textbf{53.7\%}} & 
 85.5\% & 92.3\% & 98.1\% &
7.1 & 10.3 & 145.0\% & 1 & $\sim$1.2\\

\texttt{EBH}~\cite{innamorati2018learning} & 
53.2\% & 
\emph{\textbf{85.9\%}} & \emph{\textbf{92.6\%}} & \emph{\textbf{98.3\%}} &
\emph{\textbf{6.8}} & 9.3 & 135.9\% & 9 & $\sim$12.5\\

\texttt{Diff} (ours) &  
\textbf{55.8}\% & 
\textbf{86.1\%} & \textbf{92.8\%} & \textbf{98.4\%} &
\textbf{6.6} & \textbf{8.0} & \textbf{120.7\%} & 1 & $\sim$1.6\\

\hline
STD \texttt{Diff}$^\mathrm{d}$ &  
$\pm${0.92}\% & 
$\pm${0.10\%} & $\pm${0.06\%} & $\pm${0.02\%} &
$\pm${0.05} & $\pm${0.12} & $\pm${1.8\%} &  & \\

\hline

\multicolumn{10}{l}{$^\mathrm{a}$ The best and second-best scores in each column are respectively printed in boldface and italic-boldface.}\\
\multicolumn{10}{l}{$^\mathrm{b}$ This column shows the total number of kernels to
be trained for one target kernel; only EBH introduces eight duplicates.}\\
\multicolumn{10}{l}{$^\mathrm{c}$ This column shows the approximate wall-time (relative to \texttt{Zero}) required to train a U-Net with input shape [64, 3, 224, 224], measured for forward }\\
\multicolumn{10}{l}{$^\mathrm{\ }$  and  back propagation on an Nvidia A100. The first four methods are PyTorch built-ins while the rest are based on our implementation.}\\
\multicolumn{10}{l}{$^\mathrm{d}$ For brevity, we only show the standard deviations of the metrics yielded by \texttt{Diff}. The metrics are stable with respect to model initialisation.}

\end{tabular}
\label{tab:metrics}
\end{center}
\end{table*}

\subsubsection{Semantic Segmentation}
\label{sec:segment}

In this experiment, we use the Cityscapes dataset~\cite{fei2006one} for end-to-end supervised learning of semantic segmentation with a U-Net. The category identities (8 classes) instead of the fine identities (34 classes) are used as the labels because our simple architecture and loss function (cross entropy) could not well handle a high degree of class imbalance in the latter. The original images  ($1024\times2048$) are decimated by a factor of two due to our device capacity. 

The segmentation metric scores on the validation set are shown in Table~\ref{tab:metrics}. It can be seen that the scores from \texttt{Zero}, \texttt{Refl}, \texttt{Repl}, \texttt{Circ} and \texttt{Rand} are mostly identical, implying that none of them have facilitated segmentation from the perspective of boundary handling. \texttt{Extr} and \texttt{Part} have led to a small (yet visible) improvement, and \texttt{EBH} has advanced further. Our method \texttt{Diff} has yielded the best results. Note that all the methods have attained a high baseline ($\text{accuracy}>97\%$) in an absolute sense, from which even a small improvement is not easy to achieve.

\subsubsection{Super-resolution Reconstruction}
\label{sec:super}
\begin{figure*}
    \centering
    \includegraphics[width=\textwidth]{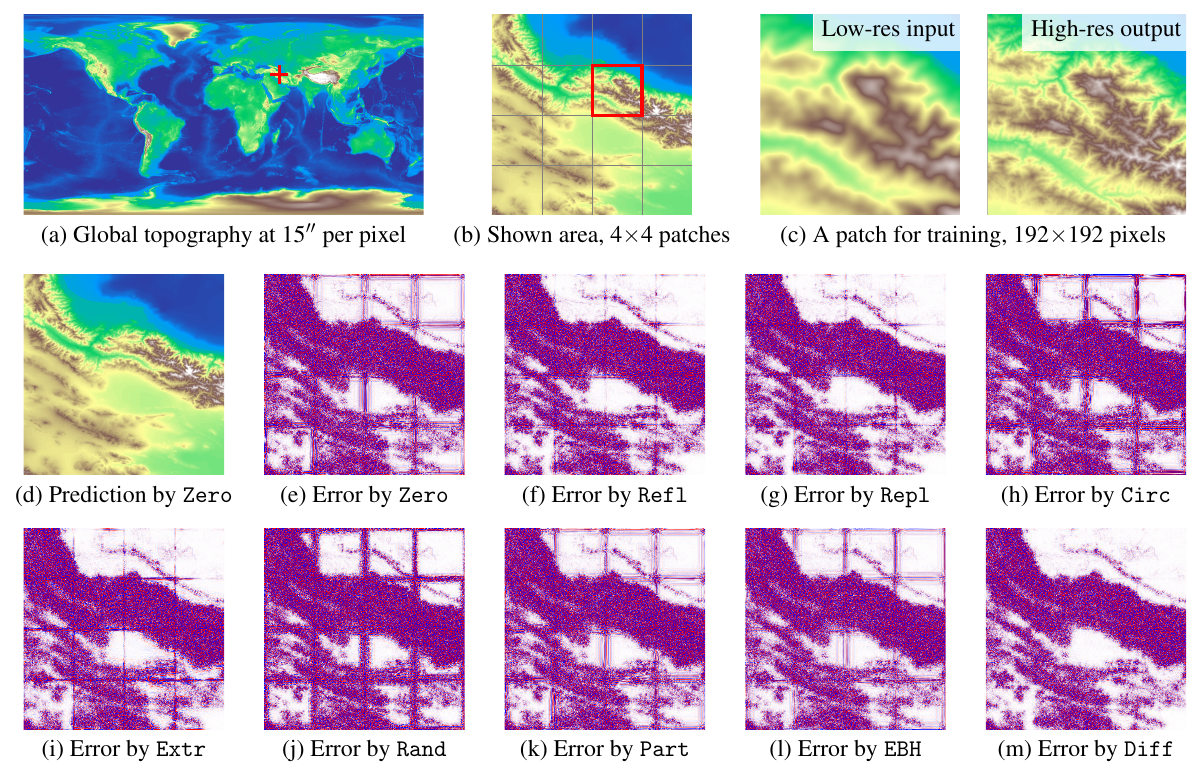}
    \caption{Super-resolution reconstruction of a world topographic map using a U-Net. The top row shows the data at three scales, (a) for global, (b) zooming into the region marked by the red-coloured cross in (a), and (c) zooming into the red-coloured box in (b). (c) shows a pair of input and output of the U-Net.
    (d) shows the reconstruction of (b) by model \texttt{Zero}; the reconstructions by the other models look similar. (e)$\sim$(m) display the error maps over (b), obtained by patch reconstruction, assembling the $4\times4$ error maps (non-overlapping), and the Farid transform (both horizontal and vertical) to highlight the artefacts near the patch boundaries (the horizontal and vertical stripes).}
    \label{fig:etopo}
\end{figure*}

In this experiment, we train a U-Net to reconstruct a world topographic map from a low to a high resolution. The data come from the \emph{ETOPO 2022 15 Arc-Second Global Relief Model}~\cite{noaa2023etopo}, a large image containing $43200\times86400$ pixels, each spanning a central angle of $15''$ (or 0.464~km on Earth's surface) in the latitudinal and longitudinal directions, as displayed in Fig.~\ref{fig:etopo}a and \ref{fig:etopo}b. We train the U-Net with non-overlapping small patches sampled from this large image, each with size $192\times 192$ (geographically $0.8^\circ \times 0.8^\circ$). We generate the low-resolution input by a Gaussian filter ($\sigma=3$), from which we attempt to recover the high-resolution output, as shown in Fig.~\ref{fig:etopo}c.
Mean squared error (MSE, denoted $\varepsilon^2$) is used as the loss function. 

To make the boundary effects more visible, we divide each test patch into two parts, interior and frame, with a frame width of eight pixels. The reconstruction error is computed separately over these two parts. For our patches of size $192\times192$, $\varepsilon^2_\text{INTER}$ is computed over the central part of size $176\times 176$ ($176=192-2\times8$), and $\varepsilon^2_\text{FRAME}$ over the cropped frame.
The MSEs are summarised in Table~\ref{tab:metrics}, which show that our method (\texttt{Diff}) has not only achieved the highest accuracy for both interior and frame but also maximally reduced the error gap between interior and frame. We visualise the error maps over a randomly picked region (near Caspian Sea), as shown in Fig.~\ref{fig:etopo}e$\sim$\ref{fig:etopo}m. 
These error maps are obtained in three steps: patch reconstruction by the U-Net, assembling the $4\times 4$ non-overlapping error maps, and applying the Farid transform~\cite{farid2004differentiation} to detect the horizontal and vertical edges. It is shown that the boundary artefacts are visible in all the error maps except the one delivered by \texttt{Diff}. 
Table~\ref{tab:metrics} and Fig.~\ref{fig:etopo} make it evident that \texttt{Diff} performs significantly better than the other methods for this super-resolution task.

\section{Conclusions}
\label{sec:con}
We have presented a new padding-free method for size-keeping image convolution. The central idea is to establish an equivalence between window-wise convolution over a discrete image and pixel-wise differentiation over a continuous representation of that image. Convolution within an incomplete sliding window can then be achieved by differentiation at its centre, with the continuous image parameterised from the nearest complete window. As such, our method preserves the differential characteristics of kernels. Our final formula is simple and computationally lightweight, available for both image filtering and CNN-based machine learning. The preservation of the differential operator at the boundary pixels makes our method more accurate for processing images with smoother boundaries, such as mathematical or physical fields and high resolution images. Our experiments have shown visible superiority of our method on both image filtering and CNN-based computer vision tasks. We provide an optimised implementation of our method, including both forward filtering (to replace \texttt{torch.nn.functional.conv2d}) and a convolutional layer class (to replace \texttt{torch.nn.Conv2d}), available open-source from \url{https://github.com/stfc-sciml/DifferentialConv2d} (with all experiments included).

\section*{Acknowledgements}
This work is supported by the EPSRC grant, Blueprinting for AI for Science at Exascale (BASE-II, EP/X019918/1), which is Phase~II of the Benchmarking for AI for Science at Exascale (BASE) grant.

\bibliographystyle{IEEEtran}
\bibliography{egbib}

\end{document}